\documentclass{article}
\usepackage{spconf,amsmath,epsfig}

\usepackage{tabularx}
\usepackage{arydshln}

\usepackage[inline,shortlabels]{enumitem}

\usepackage{siunitx}

\usepackage{tikz}
\usepackage{pgfplots}

\usepackage[nolist,nohyperlinks]{acronym}

\usepackage{cite}

\usepackage{etoolbox}
\apptocmd{\sloppy}{\hbadness 10000\relax}{}{}

\usepackage{silence}
\usepackage[skip=0.5em]{caption}

\PassOptionsToPackage{hyphens}{url}\usepackage[hidelinks]{hyperref}

\newcolumntype{C}{>{\centering\arraybackslash}X}

\DeclareSIUnit{\pixel}{pixels}
\DeclareSIUnit{\band}{bands}
\DeclareSIUnit{\sband}{spectral \ bands}
\DeclareSIUnit{\epoch}{epochs}
\DeclareSIUnit{\megapixel}{MP}
\DeclareSIUnit{\bit}{bit}
\DeclareSIUnit{\bpppc}{bpppc}
\DeclareSIUnit{\day}{days}

\pgfplotsset{
    compat=newest,
    grid=both,
    grid style={line width=.1pt, draw=gray!10},
    major grid style={line width=.2pt,draw=gray!50},
    every axis plot/.append style={line width=0.8pt},
}

\title{\uppercase{HySpecNet-11k: A LARGE-SCALE HYPERSPECTRAL DATASET FOR BENCHMARKING LEARNING-BASED HYPERSPECTRAL IMAGE COMPRESSION METHODS}}
\name{Martin Hermann Paul Fuchs $^{1}$, Begüm Demir $^{1,2}$}
\address{
    $^1$Faculty of Electrical Engineering and Computer Science, Technische Universität Berlin, Germany\\%
    $^2$BIFOLD - Berlin Institute for the Foundations of Learning and Data, Germany
}
\begin{document}
\begin{acronym}
    \acro{hyspecnet}[HySpecNet]{Hyperspectral Net}
    \acro{hyspecnet11k}[HySpecNet-11k]{Hyperspectral Net 11k}
    
    \acro{enmap}[EnMAP]{Environmental Mapping and Analysis Program}
    \acro{l2a}[L2A]{Level 2A}

    \acro{gpu}[GPU]{graphics processing unit}
    \acro{rgb}[RGB]{red, green and blue}
    \acro{hsi}[HSI]{hyperspectral image}
    \acro{cnn}[convolutional neural network]{CNN}
    
    \acro{psnr}[PSNR]{peak signal-to-noise ratio}
    \acro{mse}[MSE]{mean squared error}
    \acro{ssim}[SSIM]{structural similarity index measure}
    \acro{cr}[CR]{compression ratio}

    \acro{bpppc}[\si{\bpppc}]{bits per pixel per channel}
    \acro{decibel}[\si{\decibel}]{decibels}
    \acro{megapixel}[\si{\megapixel}]{megapixel}
    \acro{gsd}[GSD]{ground sample distance}

    \acro{cae}[CAE]{convolutional autoencoder}

    \acro{leakyrelu}[LeakyReLU]{leaky rectified linear unit}
    \acro{prelu}[PReLU]{parametric rectified linear unit}
    \acro{hardtanh}[HardTanh]{hard hyperbolic tangent}
    
    \acro{1dcae}[1D-CAE]{1D-Convolutional Autoencoder}
    \acro{adv1dcae}[1D-CAE-Adv]{Advanced 1D-Convolutional Autoencoder}
    \acro{ext1dcae}[1D-CAE-Ext]{Extended 1D-Convolutional Autoencoder}
    \acro{sscnet}[SSCNet]{Spectral Signals Compressor Network}
    \acro{3dcae}[3D-CAE]{3D Convolutional Auto-Encoder}
\end{acronym}
\maketitle
\begin{abstract}
The development of learning-based hyperspectral image compression methods has recently attracted great attention in remote sensing. Such methods require a high number of hyperspectral images to be used during training to optimize all parameters and reach a high compression performance.
However, existing hyperspectral datasets are not sufficient to train and evaluate learning-based compression methods, which hinders the research in this field.
To address this problem, in this paper we present HySpecNet-11k that is a large-scale hyperspectral benchmark dataset made up of \num{11483} nonoverlapping image patches. Each patch is a portion of \qtyproduct{128x128}{} \si{\pixel} with \SI{224}{\sband} and a \acl{gsd} of \SI{30}{\meter}.
We exploit HySpecNet-11k to benchmark the current state of the art in learning-based \acl{hsi} compression by focussing our attention on various 1D, 2D and 3D \acl{cae} architectures.
Nevertheless, HySpecNet-11k can be used for any unsupervised learning task in the framework of hyperspectral image analysis.
The dataset, our code and the pre-trained weights are publicly available at \url{https://hyspecnet.rsim.berlin}.
\end{abstract}
\begin{keywords}
EnMAP, hyperspectral dataset, image compression, deep learning, remote sensing.
\end{keywords}
\section{Introduction}
\label{sec:introduction}
Advancements in hyperspectral imaging technologies have led to a significant increase in the volume of hyperspectral data archives \cite{landgrebe2002hyperspectral}. Dense spectral information provided by hyperspectral images leads to a very high capability for the identification and discrimination of the materials in a given scene. However, to reduce the storage required for the huge amounts of hyperspectral data, it is needed to compress the images before storing them. Accordingly, one emerging research topic is associated to the efficient and effective compression of \acp{hsi} \cite{dua2020comprehensive}.

Many \ac{hsi} compression algorithms are presented in the literature. Generally, they can be divided into two categories:
\begin{enumerate*}[i)] %
    \item traditional methods, e.g. \cite{lim2001compression,penna2006new,du2007hyperspectral}; and
    \item learning-based methods, e.g. \cite{kuester20211d, kuester2022transferability, kuester2022investigating, la2022hyperspectral, chong2021end}.
\end{enumerate*} %
The most popular traditional algorithms are defined based on transform coding in combination with a quantization step and entropy coding. In contrast, learning-based methods mostly rely on \acp{cae} to reduce the dimensionality of the latent space. Recent studies show that learning-based \ac{hsi} compression methods can preserve the reconstruction quality at lower rates compared to traditional compression approaches.

Learning-based compression methods generally require a large number of unlabeled images to optimize their model parameters during training. There are only few hyperspectral benchmark datasets publicly available in remote sensing (see \autoref{tab:dataset-comparison}). To the best of our knowledge, most of the existing datasets only contain a single \ac{hsi}, which is divided into patches for the training and evaluation processes. Thus, they are not sufficient to train learning-based compression methods to reach a high generalization ability as the models may overfit dramatically, when using such training data from spatially joint areas. The lack of a large hyperspectral dataset is an important bottleneck, affecting the research and development in the field of learning-based \ac{hsi} compression. Hence, a large-scale benchmark archive consisting of a high number of \acp{hsi} acquired in spatially disjoint geographical areas is needed.

\begin{table*}
    \centering
    \caption{A summary of publicly available hyperspectral benchmark datasets and their characteristics.}
    \begin{tabularx}{\linewidth}{|C|c|c|c|c|c|c|}
        \hline
        Dataset & Acquisition & Sensor & \acs{gsd} & Spectral Range & \#Bands & Dataset Size \\ 
        \hline
        Indian Pines & 1992 & AVIRIS & \SI{20.0}{\meter} & \SIrange{400}{2500}{\nano\meter} & \num{224} & \phantom{00}\SI[round-mode=places,round-precision=2]{0.021025}{\megapixel} \\ 
        Kennedy Space Center (KSC) & 1996 & AVIRIS & \SI{18.0}{\meter} & \SIrange{400}{2500}{\nano\meter} & \num{224} & \phantom{00}\SI[round-mode=places,round-precision=2]{0.314368}{\megapixel} \\ 
        Salinas Scene & 1998 & AVIRIS & \phantom{0}\SI{3.7}{\meter} & \SIrange{420}{2450}{\nano\meter} & \num{224} & \phantom{00}\SI[round-mode=places,round-precision=2]{0.111104}{\megapixel} \\ 
        Pavia Center & 2001 & ROSIS & \phantom{0}\SI{1.3}{\meter} & \num{430}\phantom{$\,$}--\phantom{$\,$0}\SI{860}{\nano\meter} & \num{102} & \phantom{00}\SI[round-mode=places,round-precision=2]{1.201216}{\megapixel} \\ 
        Pavia University & 2001 & ROSIS & \phantom{0}\SI{1.3}{\meter} & \num{430}\phantom{$\,$}--\phantom{$\,$0}\SI{860}{\nano\meter} & \num{103} & \phantom{00}\SI[round-mode=places,round-precision=2]{0.2074}{\megapixel} \\ 
        Botswana & 2001 & Hyperion & \SI{30.0}{\meter} & \SIrange{400}{2500}{\nano\meter} & \num{242} & \phantom{00}\SI[round-mode=places,round-precision=2]{0.377856}{\megapixel} \\ 
        Cooke City & 2008 & HyMap & \phantom{0}\SI{3.0}{\meter} & \SIrange{450}{2480}{\nano\meter} & \num{126} & \phantom{00}\SI[round-mode=places,round-precision=2]{0.224}{\megapixel} \\ 
        ShanDongFeiCheng (SDFC) & 2021 & HAHS & \phantom{0}\SI{0.5}{\meter} & \SIrange{400}{1000}{\nano\meter} & \phantom{0}\num{63} & \phantom{00}\SI[round-mode=places,round-precision=2]{0.721801}{\megapixel} \\ 
        \hdashline
        HySpecNet-11k (ours) & \num{2022} & \acs{enmap} & \SI{30.0}{\meter} & \SIrange{420}{2450}{\nano\meter} & \num{224} & \SI[round-mode=places,round-precision=2]{188.137472}{\megapixel} \\ 
        \hline
    \end{tabularx}
    \label{tab:dataset-comparison}
\end{table*}

To address this problem, in this paper we introduce a new hyperspectral benchmark dataset (denoted as HySpecNet-11k) and exploit it to benchmark the current state of the art in learning-based \ac{hsi} compression.

\section{The HySpecNet-11k Dataset}
\label{sec:dataset}
HySpecNet-11k is made up of \num{11483} image patches acquired by the \acf{enmap} satellite \cite{guanter2015enmap}. Each image patch in HySpecNet-11k consists of \qtyproduct{128x128}{} \si{\pixel} and \SI{224}{\band} with a \ac{gsd} of \SI{30}{\meter} (see \autoref{tab:dataset-comparison}).

To construct HySpecNet-11k, a total of \num{250} \ac{enmap} tiles acquired during the routine operation phase between 2 November 2022 and 9 November 2022 were considered. It is worth nothing that the considered tiles are associated with less than \SI{10}{\percent} cloud and snow cover.
The tiles were radiometrically, geometrically and atmospherically corrected (\acs{l2a} water \& land product). Then, the tiles were divided into nonoverlapping image patches. Therefore, cropped patches at the borders of the tiles were eliminated. Thus, we were able to generate more than \num{45} patches per tile resulting in an overall number of \num{11483} patches for the full dataset.
Due to the \acs{l2a}-processed data, the number of bands is reduced from \num{224} to \num{202} by removing bands [\SIrange{127}{141}{}] and [\SIrange{161}{167}{}] that are affected by strong water vapor absorption.

We provide predefined splits to make the results of the considered methods reproducible.
Therefore, we randomly divided the dataset into:
\begin{enumerate*}[i)] %
   \item a training set that includes \SI{70}{\percent} of the patches,
   \item a validation set that includes \SI{20}{\percent} of the patches, and
   \item a test set that includes \SI{10}{\percent} of the patches.
\end{enumerate*} %
Depending on the way that we used for splitting the dataset, we define two different splits:
\begin{enumerate*}[i)] %
   \item an easy split, where patches from the same tile can be present in different sets (patchwise splitting); and
   \item a hard split, where all patches from one tile must belong to the same set (tilewise splitting).
\end{enumerate*} %

\begin{figure*}
    \begin{minipage}[b]{.118\linewidth}
        \centering
        \includegraphics[width=\linewidth]{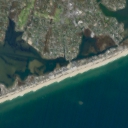}
    \end{minipage}
    \hfill
    \begin{minipage}[b]{.118\linewidth}
        \centering
        \includegraphics[width=\linewidth]{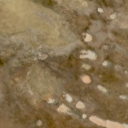}
    \end{minipage}
    \hfill
    \begin{minipage}[b]{.118\linewidth}
        \centering
        \includegraphics[width=\linewidth]{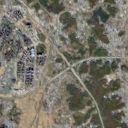}
    \end{minipage}
    \hfill
    \begin{minipage}[b]{.118\linewidth}
        \centering
        \includegraphics[width=\linewidth]{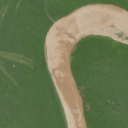}
    \end{minipage}
    \hfill
    \begin{minipage}[b]{.118\linewidth}
        \centering
        \includegraphics[width=\linewidth]{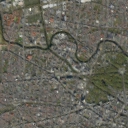}
    \end{minipage}
    \hfill
    \begin{minipage}[b]{.118\linewidth}
        \centering
        \includegraphics[width=\linewidth]{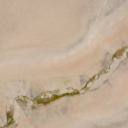}
    \end{minipage}
    \hfill
    \begin{minipage}[b]{.118\linewidth}
        \centering
        \includegraphics[width=\linewidth]{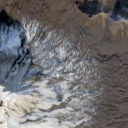}
    \end{minipage}
    \hfill
    \begin{minipage}[b]{.118\linewidth}
        \centering
        \includegraphics[width=\linewidth]{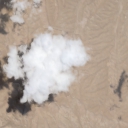}
    \end{minipage}
    \begin{minipage}[b]{.118\linewidth}
        \vspace{.04\linewidth}
        \centering
        \includegraphics[width=\linewidth]{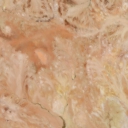}
    \end{minipage}
    \hfill
    \begin{minipage}[b]{.118\linewidth}
        \centering
        \includegraphics[width=\linewidth]{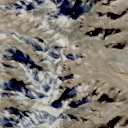}
    \end{minipage}
    \hfill
    \begin{minipage}[b]{.118\linewidth}
        \centering
        \includegraphics[width=\linewidth]{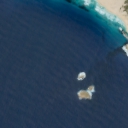}
    \end{minipage}
    \hfill
    \begin{minipage}[b]{.118\linewidth}
        \centering
        \includegraphics[width=\linewidth]{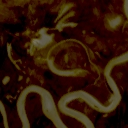}
    \end{minipage}
    \hfill
    \begin{minipage}[b]{.118\linewidth}
        \centering
        \includegraphics[width=\linewidth]{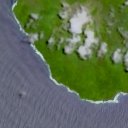}
    \end{minipage}
    \hfill
    \begin{minipage}[b]{.118\linewidth}
        \centering
        \includegraphics[width=\linewidth]{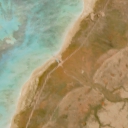}
    \end{minipage}
    \hfill
    \begin{minipage}[b]{.118\linewidth}
        \centering
        \includegraphics[width=\linewidth]{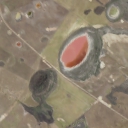}
    \end{minipage}
    \hfill
    \begin{minipage}[b]{.118\linewidth}
        \centering
        \includegraphics[width=\linewidth]{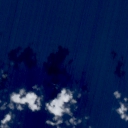}
    \end{minipage}
    \caption{True color representations of example images from our proposed HySpecNet-11k dataset. Red, green and blue channels are extracted from \ac{enmap} bands \num{43}, \num{28} and \num{10} at wavelengths \SI{634.919}{\nano\meter}, \SI{550.525}{\nano\meter} and \SI{463.584}{\nano\meter}, respectively.}
    \label{fig:exemplary_images}
\end{figure*}

To get an overview of the dataset, \autoref{fig:exemplary_images} illustrates representative HySpecNet-11k images patches.
It is worth noting that compression methods generally do not require labeled training images and therefore our dataset does not contain image annotations.

\section{Learning-based Hyperspectral Image Compression}
\label{sec:related_works}
In order to provide an initial benchmarking of the proposed HySpecNet-11k dataset in the framework of learning-based \ac{hsi} compression, we train and evaluate the following state of the art baseline methods:
\begin{enumerate*}[i)] %
    \item \ac{1dcae} \cite{kuester20211d};
    \item \ac{adv1dcae} \cite{kuester2022transferability};
    \item \ac{ext1dcae} \cite{kuester2022investigating};
    \item \ac{sscnet} \cite{la2022hyperspectral}; and
    \item \ac{3dcae} \cite{chong2021end}.
\end{enumerate*} %
All methods are based on convolutional operations paired with down- and upsamplings in the encoder and decoder, respectively. They differ from each other with respect to the approaches considered for spatial and spectral compression.

An overview of the considered methods is given in the following. For a detailed description we refer the reader to the respective papers.

The \acf{1dcae} \cite{kuester20211d} applies pixelwise compression without utilizing spatial content. The encoder consists of multiple 1D convolutions in combination with two 1D max pooling layers that fix the \ac{cr} to a factor of \num{4}. \Acp{leakyrelu} are present as activation functions after each convolution.
The decoder mirrors the encoder, however it uses two upsampling layers to reverse the respective downsamplings from the encoder. A sigmoid activation function scales the reconstructed pixel values to the range from \SIrange{0}{1}{} in the last decoder layer. Furthermore, padding and unpadding are needed in encoder and decoder, respectively.
The \acf{adv1dcae} \cite{kuester2022transferability} adapts the decoder by substituting the combinations of 1D convolutional layer and upsampling layer with 1D transposed convolutions to make the upsampling operation trainable and more adaptive. On the encoder side, the 1D max poolings are dropped for 1D average poolings.
Moreover, the \acf{ext1dcae} \cite{kuester2022investigating} adds a 1D batch normalization \cite{ioffe2015batch} between each (transposed) convolutional and activation layer. The sigmoid activation in the last decoder layer is replaced by a \ac{hardtanh} and 1D average poolings are substituted by 1D max poolings again. The repetition of down- and upsampling blocks enables the construction of models with different \acp{cr}.

The \acf{sscnet} \cite{la2022hyperspectral} encoder uses 2D convolutions with \acp{prelu} as activation after each convolutional layer. Three 2D max pooling layers are added for a fixed spatial compression by a factor of \num{64}. The final \ac{cr} is set via the number of latent channels in the bottleneck layer.  Simultaneously, the ratio between input and latent channels decides the spectral compression factor.
2D transposed convolutions are used to reconstruct the \acp{hsi} on the decoder side. After each transposed convolution there is a \ac{prelu} activation except for the last layer where a sigmoid activation is used for scaling the outputs into the correct range from \SIrange{0}{1}{}. For all convolutional layers a kernel size of \qtyproduct{3x3}{} \si{\pixel} ensures the integration of nearby spatial content into the compression of the currently considered pixel.

The \ac{3dcae} \cite{chong2021end} is built by three strided 3D convolutional layers, each of which is followed by a 3D batch normalization \cite{ioffe2015batch} and a \ac{leakyrelu}. Residual blocks are added to increase the depth of the network. Upsampling layers are furthermore present on the decoder side for reconstruction. Padding operations are required as in the \ac{1dcae}. The 3D kernels in the convolutional layers are able to integrate the local  spatial and spectral neighborhood jointly during the compression. Spatial and spectral \acp{cr} are fixed to a factor of \num{64} and \num{4}, respectively and the overall \ac{cr} can be set via the number of latent channels in the bottleneck of the network. 

\section{Experimental Results}
\label{sec:experimental_results}
Our code is implemented in PyTorch based on the CompressAI \cite{begaint2020compressai} framework.
We applied gradient clipping and a global min-max normalization to scale the input data in the range between \SIrange{0}{1}{} that is usually a requirement for learning-based compression methods.
As optimizer, we used Adam \cite{kingma2014adam} with a learning rate of \num{1e-4} for the \ac{1dcae} \cite{kuester20211d}, \ac{adv1dcae} \cite{kuester2022transferability} and \ac{ext1dcae} \cite{kuester2022investigating}. For \ac{sscnet} \cite{la2022hyperspectral} and \ac{3dcae} \cite{chong2021end} the learning rate was set to \num{1e-5}. We trained the networks using \ac{mse} as loss function until convergence on the validation set that took \SI{500}{\epoch}, \SI{2000}{\epoch} and \SI{1000}{\epoch} for the \acp{1dcae}, \ac{sscnet} and \ac{3dcae}, respectively.
Training runs were carried out on a single NVIDIA A100 SXM4 80 GB \acs{gpu} and took between \SIrange{1}{10}{\day} each, depending on the method.
In general, \ac{sscnet} requires fewer \acs{gpu} hours than \ac{1dcae} and \ac{3dcae}.
For the \ac{1dcae} another factor is the \ac{cr}. The higher the \ac{cr}, the deeper the network and thus the longer the training time. Therefore, the modified versions of the \ac{1dcae} with increased \acp{cr} were trained for only \SI{250}{\epoch} due to runtime limitations.

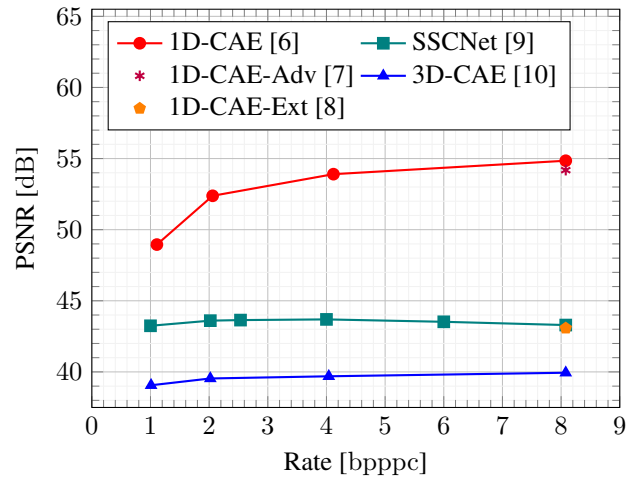
\begin{figure}
    \centering
    \begin{tikzpicture}
        \begin{axis}[
                xlabel={Rate [\si{\bpppc}]},
                ylabel={PSNR [\si{\decibel}]},
                xmin=0,xmax=9,
                xtick={0,1,...,9},
                ymin=37.5,ymax=65.5,
                minor tick num=4,
                legend pos=north west,
                width=\linewidth,
                height=0.8\linewidth,
                legend columns=2,
                legend cell align={left},
            ]
            \addplot[red,mark=*] coordinates {
                (1.1089108910891081, 48.952538482825375)    
                (2.059405940594064, 52.38174220872962)      
                (4.118811881188128, 53.902654146940)        
                (8.079207920792062, 54.84735990895165)      
            };
            \addlegendentry{\acs{1dcae} \cite{kuester20211d}};

            \addplot[teal,mark=square*] coordinates {
                (1.0, 43.2428080505795)                     
                (2.0198019801980154, 43.60347966353098)     
                (2.5346534653465365, 43.64347089661492)     
                (4.0, 43.69138220945994)                    
                (6.0, 43.52298931280772)                    
                (8.079207920792062, 43.29476347234514)      
            };
            \addlegendentry{\acs{sscnet} \cite{la2022hyperspectral}};
            
            \addplot[purple,mark=asterisk,only marks] coordinates {
                (8.079207920792062, 54.184891091452705)     
            };
            \addlegendentry{\acs{adv1dcae} \cite{kuester2022transferability}};

            \addplot[blue,mark=triangle*] coordinates {
                (1.0099009900990086, 39.06108974227706)     
                (2.0198019801980154, 39.5384429163403)      
                (4.039603960396031, 39.6920463376575)       
                (8.079207920792062, 39.94142961502075)      
            };
            \addlegendentry{\acs{3dcae} \cite{chong2021end}};

            \addplot[orange,mark=pentagon*,only marks] coordinates {
                (8.079207920792062, 43.07850521140628)      
            };
            \addlegendentry{\acs{ext1dcae} \cite{kuester2022investigating}};
        \end{axis}
    \end{tikzpicture}
    \caption{Rate-distortion performance of learning-based \acl{hsi} compression methods on the test set of our proposed HySpecNet-11k dataset (easy split). Rate is visualized in \acf{bpppc} and distortion is given as \acf{psnr} in \acf{decibel}.}
    \label{fig:rate-distortion-plot}
\end{figure}

Rate-distortion performance of the baseline methods on the HySpecNet-11k test set is shown in \autoref{fig:rate-distortion-plot}. The results have been obtained by using the HySpecNet-11k easy split as introduced in \autoref{sec:dataset}.
Overall, the \ac{1dcae} produces the highest quality reconstructions with a \ac{psnr} of \SI[round-mode=places,round-precision=2]{54.84735990895165}{\decibel} at a relatively high fixed rate of \SI[round-mode=places,round-precision=2]{8.079207921}{\bpppc}.
Despite the added trainable upsampling operations, the \ac{adv1dcae} reaches a \SI[round-mode=places,round-precision=2]{0.662468817}{\decibel} lower \ac{psnr} than the \ac{1dcae} at the same rate and the \ac{ext1dcae} only achieves a \ac{psnr} of \SI[round-mode=places,round-precision=2]{43.07850521140628}{\decibel}, while increasing training time significantly due to batch normalization. As a result, further experiments were only performed for the \ac{1dcae}.
\ac{sscnet} is able to operate at a lower rate of \SI[round-mode=places,round-precision=2]{2.5346534653465365}{\bpppc} due to spatial dimensionality reduction, which on the other hand introduces blurry reconstructions that reduce the \ac{psnr} to \SI[round-mode=places,round-precision=2]{43.64347089661492}{\decibel}. 
The \ac{3dcae} only achieves \SI[round-mode=places,round-precision=2]{39.5384429163403}{\decibel} \ac{psnr} at \SI[round-mode=places,round-precision=2]{2.0198019801980154}{\bpppc} while being particularly unstable on the validation set during training.
We would like to note that we have observed similar behaviour on the HySpecNet-11k hard split, which shows the high generalization ability of the considered methods due to our proposed large-scale dataset.

To achieve a better comparison along several \acp{cr}, we modified the baseline models by repeating the downsampling blocks in the \ac{1dcae} and varying the number of latent channels for \ac{sscnet} and \ac{3dcae}.
As seen in \autoref{fig:rate-distortion-plot}, the \ac{1dcae}, which only compresses the spectral content, is superior in all cases compared to both other methods that apply spatial downsampling.
Even when increasing the number of latent channels in the bottleneck, \ac{sscnet} and \ac{3dcae} are not able to compensate the loss of information introduced by the spatial compression.
Thus, in contrast to \acs{rgb} imagery, \ac{hsi} compression methods should focus implicitly on the spectral domain while maintaining the spatial dimensions when applied on hyperspectral data with a low spatial resolution.

\section{Conclusion}
\label{sec:conclusion}
In this paper, we have introduced HySpecNet-11k that is a large-scale hyperspectral benchmark dataset (which consist of \num{11483} unlabeled image patches) for learning-based hyperspectral image compression problems. To the best of our knowledge, HySpecNet-11k is the first publicly available benchmark dataset that includes images acquired by the \acs{enmap} satellite.
It is worth nothing that the use of HySpecNet-11k is not limited to image compression problems and can be exploited for any unsupervised learning task. We believe that HySpecNet-11k will make a significant advancement in the field of unsupervised learning from hyperspectral data by overcoming the current limitations of existing \acl{hsi} datasets.
With the continuous release of \acs{enmap} tiles we plan to enrich the dataset and develop further extended versions of HySpecNet as a future work.

\section{Acknowledgements}
This work is supported by the European Research Council (ERC) through the ERC-2017-STG BigEarth Project under Grant 759764.

\bibliographystyle{IEEEtran}
\bibliography{refs,refs-datasets,refs-traditional_methods,refs-learning_methods}

\end{document}